\documentclass[runningheads]{llncs}

\usepackage{eccv}

\usepackage{eccvabbrv}

\usepackage{graphicx}
\usepackage{booktabs}

\usepackage{multirow}
\usepackage{algorithm}
\usepackage{algorithmic}
\usepackage{caption} 
\usepackage{amssymb}
\usepackage{blindtext}

\usepackage[accsupp]{axessibility}  

\usepackage{hyperref}

\usepackage{orcidlink}

\begin{document}

\newcommand{\methodname}{{FlexAttention}}
\title{{\methodname} for Efficient High-Resolution Vision-Language Models}


\author{
Junyan Li\inst{1}\orcidlink{0009-0005-3557-4922} \and
Delin Chen\inst{1}\orcidlink{0000-0002-9519-093X} \and
Tianle Cai\inst{2}\orcidlink{0009-0008-8644-3895} \and
Peihao Chen\inst{3}\orcidlink{0000-0002-6847-1621} \and
Yining Hong\inst{4}\orcidlink{0000-0002-0518-2099} \and
Zhenfang Chen\inst{5}\orcidlink{0009-0005-1619-932X} \and
Yikang Shen\inst{5}\orcidlink{0000-0001-6836-0510} \and
Chuang Gan\inst{1,5}\orcidlink{0000-0003-4031-5886}
}

\authorrunning{J. Li et al.}

\institute{
UMass Amherst \and
Princeton University \and
South China University of Technology \and
University of California, Los Angeles \and
MIT-IBM Watson AI Lab \\
\email{\{junyanli,delinchen\}@umass.edu\\
tianle.cai@princeton.edu\\
\{phchencs,yninghong,chenzhenfang2013,yikang.shn,ganchuang1990\}@gmail.com}
}

\maketitle

\begin{abstract}

\setcounter{footnote}{0}
Current high-resolution vision-language models encode images as high-resolution image tokens and exhaustively take all these tokens to compute attention, which significantly increases the computational cost. To address this problem, we propose \textsc{{\methodname}}, a flexible attention mechanism for efficient high-resolution vision-language models. Specifically, a high-resolution image is encoded both as high-resolution tokens and low-resolution tokens, where only the low-resolution tokens and a few selected high-resolution tokens are utilized to calculate the attention map, which greatly shrinks the computational cost. The high-resolution tokens are selected via a high-resolution selection module which could retrieve tokens of relevant regions based on an input attention map. The selected high-resolution tokens are then concatenated to the low-resolution tokens and text tokens, and input to a hierarchical self-attention layer which produces an attention map that could be used for the next-step high-resolution token selection. The hierarchical self-attention process and high-resolution token selection process are performed iteratively for each attention layer. Experiments on multimodal benchmarks prove that our \textsc{{\methodname}} outperforms existing high-resolution VLMs (\textit{e.g.,} relatively $\sim$9\% in V* Bench, $\sim$7\% in TextVQA), while also significantly reducing the computational cost by nearly 40\%.\footnote{Project page: \url{https://vis-www.cs.umass.edu/flexattention}}
\end{abstract}

\section{Introduction}
\label{sec:intro}
\begin{figure}[t]
    \centering
    \includegraphics[width=\linewidth]{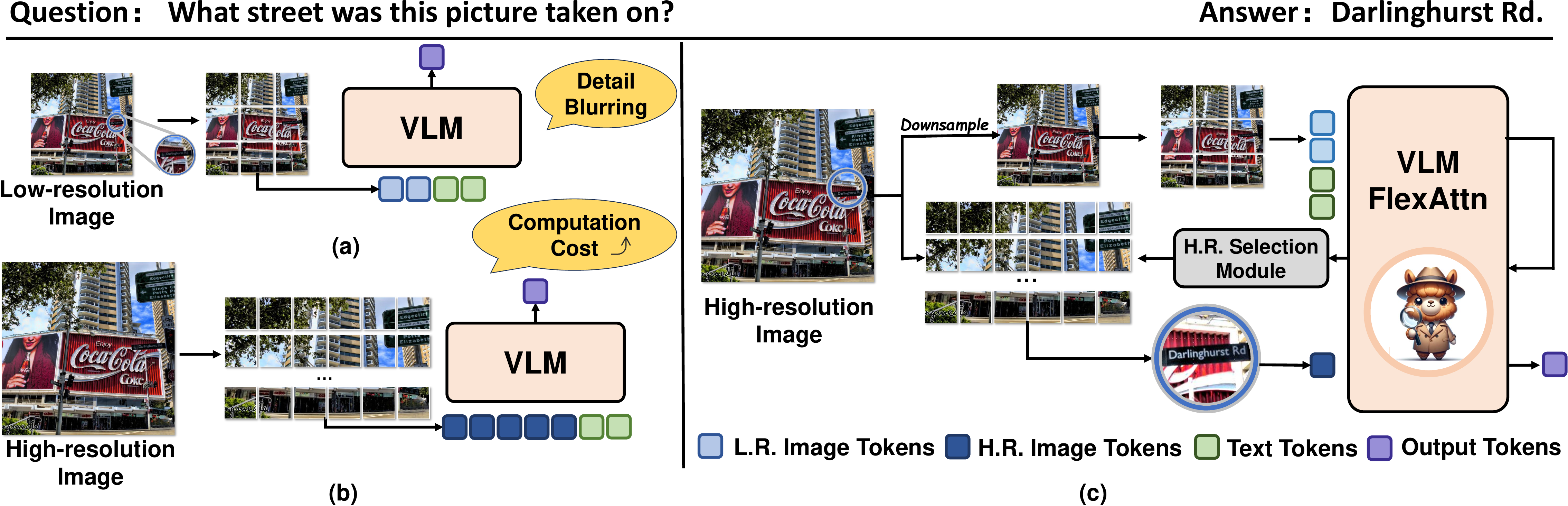}
    \caption{\textbf{An overview of VLMs processing high-resolution images for the VQA task.} \textbf{(a)} low-resolution VLM will first downsample the high-resolution image to meet its vision encoder requirement. The detail in the low-resolution image is missing, thus it is hard for it to correctly answer the question. \textbf{(b)} high-resolution VLM can take the high-resolution image as input, at the cost of a large amount of high-resolution image tokens, leading to excessive computational cost. \textbf{(c)} Equipped with our {\methodname}, the model encodes the whole high-resolution image and dynamically selects a small portion of the high-resolution feature that the model is paying attention to during the generation, thus avoiding the high computational cost.}
    \label{fig:intro}
\end{figure}

Large vision-language models (VLMs), such as those described in \cite{li2023blip2, li2023otter}, exhibit remarkable capabilities across a range of multimodal tasks including image captioning, visual question answering, image-text matching, and so on. However, these models typically process images at relatively low resolutions  (\textit{e.g.,} 224$\times$224 or 336$\times$336), thus struggling in scenarios where detailed scrutiny of small regions (e.g., minor texts or small objects) is required. This limitation becomes evident, for instance, in Fig. \ref{fig:intro} (a), where these models fail to discern the words on the printed sign due to the constraints of low-resolution inputs. 

To address this problem, several high-resolution VLMs (\textit{e.g.}, LLaVA-1.5-HD \cite{liu2023improved} and CogAgent \cite{hong2023cogagent}) have been proposed, which could take high-resolution images as inputs and encode them as high-resolution tokens. Although such models provide a more detailed examination of small regions, they exhaustively process all high-resolution tokens to compute attention, which places a heavy burden upon computational resources. These models deviate from the way human beings perform visual reasoning. Instead of memorizing all pixel-perfect details, we tend to maintain a coarse representation at first, and attend to relevant regions for retrieval of more details only when instilled with external stimuli \cite{Broadbent1958, Oishausen1993, Palmer1983, Oishausen1993}. It's essential that high-resolution VLMs could also flexibly and dynamically attend to the regions of interest based on low-resolution features for high-resolution detail retrieval.

To this end, we present \textsc{{\methodname}}, a novel attention mechanism that could be seamlessly plugged into most vision-language models to empower their abilities to perceive images with higher resolutions in an efficient manner. Specifically, as is shown in Fig. \ref{fig:intro} (c), \textsc{{\methodname}} takes a high-resolution image as input, and encodes the image both as high-resolution image tokens and low-resolution image tokens.
For computational efficiency, we only feed the low-resolution tokens and text tokens to the first few layers to roughly understand the entire image. For subsequent layers, only the low-resolution tokens and a very small portion of high-resolution tokens are utilized to calculate the attention, which significantly shrinks the computational cost. At each decoder layer with \textsc{{\methodname}}, we have a high-resolution feature selection module and a hierarchical self-attention module. The high-resolution feature selection module retrieves high-resolution tokens of relevant regions based on the input attention map. The selected high-resolution tokens are concatenated to the low-resolution tokens along with text tokens, and input to the hierarchical self-attention module. The hierarchical self-attention module produces an attention map, which could be used for the high-resolution token selection that selects high-resolution tokens that are input to the next-layer hierarchical self-attention. The two modules are iteratively processed until the last layer, which produces the final answer through a projector.

We evaluate our \textsc{{\methodname}} on several high-resolution multimodal benchmarks, including general benchmarks such as V* Bench \cite{wu2023vstarbench} and Magnifierbench \cite{li2023otterhd}, as well as domain-specific benchmarks such as TextVQA \cite{singh2019textvqa} for text understanding and RSVQA \cite{lobry2020rsvqa} for remote sensing. We show a better performance than other high-resolution methods with nearly 40\% computational cost reduction, proving the efficiency of our method. What's more, we achieve a higher score in V* Bench compared to commercial chatbots such as GPT-4V \cite{achiam2023gpt}.

\section{Related Works}
\label{sec:related}

\noindent{\textbf{Vision Language Models.}}
Our work is closely related to the research that tried to train large multimodal models~\cite{li2020oscar,zhang2021vinvl,tan2019lxmert,sun2019videobert,lu2019vilbert,zhou2020unified} for various vision language tasks like visual question answering~\cite{hudson2019gqa,singh2019textvqa}, referring expression comprehension~\cite{kazemzadeh2014referitgame,yu2016modeling} and text-based image retrieval~\cite{li2011text,young-etal-2014-image}.
Traditional methods~\cite{li2020unicoder,hao2020towards} usually collected large vision-language datasets and learned joint representation between vision and language from scratch to handle different tasks.
Such models usually worked well in in-domain data but performed inferior in the benchmarks that require common sense understanding and outside world knowledge~\cite{marino2019ok,li2023otterhd}. 

Later, large language models (LLMs)~\cite{brown2020language,touvron2023llama,team2023gemini} showed impressive power in natural language understanding and reasoning, which brought new possibilities and capabilities to the research of vision and language.  A series of large vision-language models have been proposed, which typically connect a pre-trained vision encoder~\cite{dosovitskiy2020image,radford2021learning} with a pre-trained large language model~\cite{touvron2023llama,vicuna2023}.
Flamingo~\cite{alayrac2022flamingo} first used the cross-attention mechanisms to encode the visual context into the LLMs.
BLIP2~\cite{li2022blip} proposed the QFormer, which uses a bert model~\cite{kenton2019bert} to transform the visual features into a set of learned tokens for LLMs.
Fuyu~\cite{bavishi2024fuyu} directly projected the image patches into inputs for LLMs to get rid of pre-trained vision encoders.
While these models have impressive performance on commonsense understanding and perform incredibly well on traditional vision-language tasks, they often fail to handle tasks that require high-resolution inputs~\cite{wu2023vstarbench,li2023otterhd} due to two reasons: 1) most VLMs utilize CLIP \cite{radford2021learning} as their vision encoder, and this limits the input image size of these VLMs to the fixed and relatively small resolution that CLIP is trained on (e.g., 224x224), and 2) they lack model mechanisms to efficiently handle long image patch sequences, which will lead to the excessive computational cost when the number of image patches increased quadratically with the image resolution increased.

\noindent{\textbf{High-Resolution VLMs.}}
To improve VLMs' capability to handle inputs with high resolutions, several VLMs have been proposed~\cite{hong2023cogagent,feng2023docpedia,liu2024llavanext}. DocPedia~\cite{feng2023docpedia} transformed the image into a frequency domain to maintain better semantics of the high-resolution images. LLaVA 1.6~\cite{liu2024llavanext} designed inputs of various scales to meet the needs of different tasks and balance efficiency and performance. While these models relieved the problem of dense computation, they are orthogonal to our method and have not designed any new attention mechanisms to handle the quadratic computational cost increase challenge introduced by the self-attention mechanism. Recently, CogAgent~\cite{hong2023cogagent} designed a new vision encoder for high-resolution image input. Different from us, it requires calculating the dense correspondence between the hidden states and the whole high-resolution image feature through cross-attention at every layer of the large language model, making it less efficient. Also, the data for training the model is not publicly available while we are planning to release all data, code, and models for the whole research community.

\noindent{\textbf{Efficient Mechanisms for Sequence Modeling.}}
Our work relates to the development of efficient mechanisms for sequence modeling. One approach tackles the quadratic complexity of standard attention mechanisms concerning sequence length. This is done by using structured approximations
~\cite{child2019generating,wang2020linformer, kitaev2020reformer,zaheer2020big,tay2020synthesizer,peng2021random} or linear attention \cite{cai2023efficientvit}.  Another approach replaces attention entirely with recurrent-style models, such as Recurrent Neural Networks (RNNs) and state-space models~\cite{gu2021efficiently,gu2023mamba,yang2023gated,hou2024rwkv}. Of particular relevance is the work by Yang \etal~\cite{yang2016hierarchical}, who introduced a hierarchical attention network for document classification. Their model uses a hierarchical structure and two-level attention mechanisms to improve document representation. Our work diverges by focusing on efficient mechanisms specifically designed for high-resolution image inputs, ensuring seamless cooperation with the computations of large language models.

\section{Preliminary}
\label{sec:preliminary}

\textbf{Notation.} We define some terms that will be used throughout the paper. For a high-resolution vision-language model, we define its high-resolution image input as $I_{HR}$ and the text input as $T$. Furthermore, we define the low-resolution image tokens as $f_{LR}$, the high-resolution image tokens as $f_{HR}$, and the text tokens as $f_{T}$. The hidden state for the VLM is denoted as $H \in\mathbb{R}^{N\times D}$, with a sequence length of $N$ and a hidden state size of $D$. The hidden state $H$ comprises $N_i$ low-resolution image tokens followed by $N_t$ text tokens. We define $f_{SHR}$ as the selected subset of $M$ high-resolution image tokens $f_{HR}$.

\noindent\textbf{Autoregressive Large Language Models.} Autoregressive large language models (LLMs) such as LLaMA \cite{touvron2023llama} play a crucial role in most vision-language models as they are responsible for taking both image and text tokens as input and generating the answer sequence. An autoregressive LLM is constituted by several stacked decoder layers. Each decoder layer has two sub-layers. The first is a self-attention module, and the second is a feed-forward (FFN) layer. A skip connection is employed around each of the two sub-layers, followed by layer normalization (LN). In short, the output of each sub-layer is $\text{LN}(x+\text{SubLayer}(x))$. For simplicity, layer normalization will be omitted in the subsequent discussion.

\noindent\textbf{Self-attention and Attention Map.} Self-attention \cite{vaswani2017attention} is the basic module for a decoder layer. For the self-attention, given input hidden state $H \in\mathbb{R}^{N\times D}$, it will first utilize a linear projection layer to project $H$ into $Q$, $K$, and $V$, namely the query, key, and value matrix, and performs the following calculation:

\begin{align}
    \text{Self-attention}(H) = \text{softmax}\left(\frac{QK^T}{\sqrt{d_k}}\right)V,
\end{align}

\noindent where $Q=HW_{Q}$, $K=HW_{K}$, $V=HW_{V}$ and $W_{Q}$/$W_{K}$/$W_{V}$$\in\mathbb{R}^{D\times d}$ is the learnable linear projection matrix. Specifically, the attention map $Map$ is obtained after the softmax operation:

\begin{align}
    Map = \text{softmax}\left(\frac{QK^T}{\sqrt{d_k}}\right).
\label{attention_map}
\end{align}

\noindent The attention map $Map$ is an $N$x$N$ matrix that measures the importance between tokens: the (i, j) attention value in the attention map indicates the importance of the j-th token to the i-th token, and a higher value means that the j-th token is more important to the i-th token.

\noindent\textbf{Limitation of Self-attention.} The computational cost of the self-attention mechanism is characterized by a quadratic increase relative to the sequence length $N$ of the hidden state $H$. This computational complexity is further amplified when integrating high-resolution images, as it substantially increases the number of image tokens, consequently extending the length of the hidden state. As a result, the computational requirements of the self-attention mechanism undergo a significant escalation, making the processing of high-resolution image inputs impractical due to the prohibitive computational overhead.

\section{Vision-language Model with {\methodname}}
\label{sec:method}
\begin{figure}[t]
    \centering
    \includegraphics[width=\linewidth]{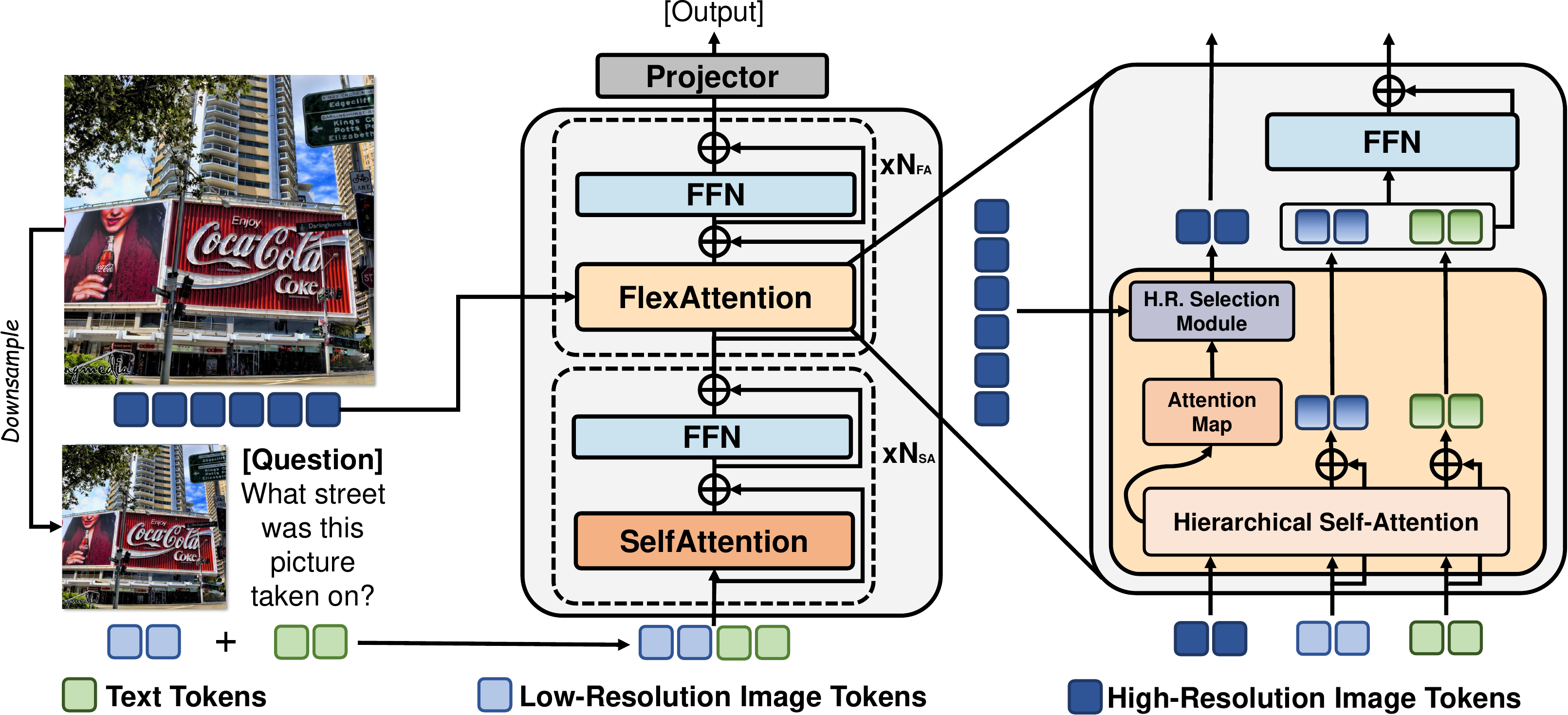}
    \caption{\textbf{An Overview of {\methodname}.} Within each \methodname~ layer, the encoded high-resolution image features are selected according to the input attention map. These selected features are then inputted into the hierarchical self-attention mechanism alongside input hidden states, which encompass both low-resolution image tokens and text tokens, for computation.}
    \label{fig:method}
\end{figure}

\subsection{Overall Architecture}

To solve the limitations of self-attention when dealing with high-resolution images, we introduce \textsc{{\methodname}}, which efficiently analyzes high-resolution images by dynamically attending to important regions of high-resolution images. The \textsc{{\methodname}}  can be plugged into most vision-language models by replacing their self-attention module with our proposed \textsc{{\methodname}} module.

As shown in Fig. \ref{fig:method}, the modified vision-language model consists of $N_{SA}+N_{FA}$ decoder layers, where the first $N_{SA}$ layers are with the vanilla self-attention module and the last $N_{FA}$ layers are with our proposed \textsc{{\methodname}} module. Given a high-resolution image, we first downsample it to a low-resolution one and feed both images into an image encoder to get high-resolution and low-resolution image tokens, respectively. For computational efficiency, we only feed the low-resolution image tokens and text tokens to the first $N_{SA}$ layers to roughly understand the whole image. For the subsequent $N_{FA}$ decoder layers with \textsc{{\methodname}}, to efficiently perceive more image details, we additionally feed it with selected high-resolution image tokens. Specifically, \textsc{{\methodname}} consists of two modules: a high-resolution feature selection module and a hierarchical self-attention module. Instead of feeding forward all high-resolution tokens, the high-resolution feature selection module flexibly selects important tokens for the next layer according to an attention map. The hierarchical self-attention module is designed to fuse the selected high-resolution information into the original hidden state. Finally, we use a projector linear layer to produce textual output.

\subsection{High-resolution Feature Selection Module}

For an autoregressive LLM, the next token is predicted by the last hidden state of the last token. By inspecting the attention values of all other tokens corresponding to the last token in the attention map in Eq. \ref{attention_map}, we can find out which tokens the model is paying attention to when generating the next predicted token. When it comes to the vision-language model, this also applies to image tokens $f_{LR}$. Those image tokens that possess a high attention value can be treated as relevant to important image regions when generating the next token. 
Although the details contained in the low-resolution image tokens are limited, we could retrieve the high-resolution details of the same image regions that have been attended to.
Therefore, instead of feeding all high-resolution tokens to the attention module which will lead to excessive computational cost, we dynamically select a very small portion (approximately 10\%) of the high-resolution tokens, namely $f_{SHR}$, and only forward this portion to the attention module.

\begin{figure}[t]
    \centering
    \includegraphics[width=\linewidth]{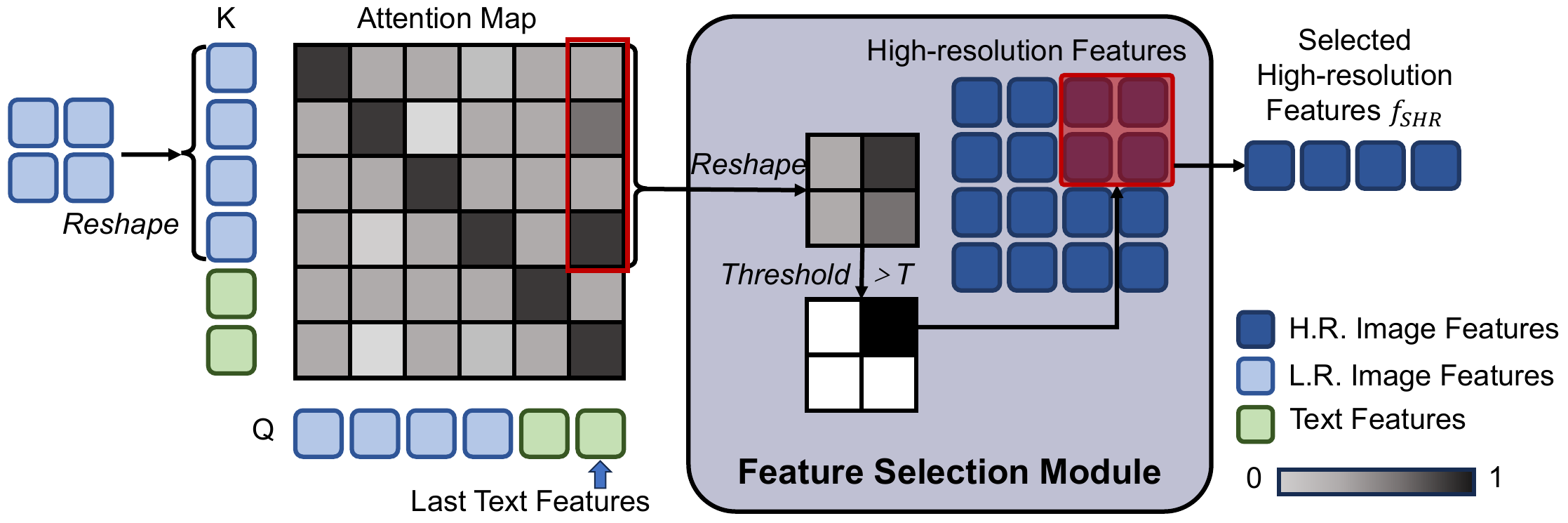}
    \caption{Illustration of high-resolution feature selection module.}
    \label{fig:selection}
\end{figure}

As is shown in Fig. \ref{fig:selection}, we take the first $N_i$ values from the last column of the attention map, which corresponds to the importance of the low-resolution image tokens to the last text token, and reshape this 1-D vector to a 2-D map, denoted as the attention mask. Each value in this mask is linked with a patch in the low-resolution image $I_{LR}$, indicating that patch's importance. The mask is normalized, binarized, and then resized to the same size as the high-resolution feature patch tokens to form the high-resolution selection mask, which serves as the selection decision on whether to select the token of a patch or not. Finally, we apply this mask to the high-resolution image tokens to get the selected high-resolution feature $f_{SHR}$.

\subsection{Hierarchical Self-attention Module}

The hierarchical self-attention is the core mechanism to fuse the information from the selected high-resolution tokens $f_{SHR}$ into the hidden state $H$ which consists of both low-resolution tokens and the text tokens. It takes the selected high-resolution tokens $f_{SHR}\in\mathbb{R}^{M\times D}$ and the hidden state $H\in\mathbb{R}^{N\times D}$ as inputs, and outputs the attention map $Map'$ and the updated hidden state $H'$. The calculation of the hierarchical self-attention is summarized as

\begin{align}
&Q = HW_Q, \\
&K_{all} = Concat(HW_K, f_{SHR}W_K'), \\
&V_{all} = Concat(HW_V, f_{SHR}W_V'), \\
&\text{Hierarchical Self-attention}(H, f_{SHR}) = \text{softmax}\left(\frac{QK_{all}^T}{\sqrt{d_k}}\right)V_{all},
\end{align}

\noindent where $W_{Q}$/$W_{K}$/$W_{V}$/$W_{K}'$/$W_{V}'$$\in\mathbb{R}^{D\times d}$ is the learnable linear projection matrix. $K_{all}\in\mathbb{R}^{(N+M)\times d}$ and $V_{all}\in\mathbb{R}^{(N+M)\times d}$ are the key and value matrix that fuses the information from high-resolution features. Similar to self-attention, we can obtain an attention map after the softmax operation: 

\begin{align}
    Map' = \text{softmax}\left(\frac{QK_{all}^T}{\sqrt{d_k}}\right).
\label{attention_map2}
\end{align}

Different from self-attention, this attention map $Map'$ has a shape of $N\times(N+M)$ as it additionally contains the attention values of high-resolution tokens corresponding to other tokens. We only keep the first $N \times N$ attention values of the matrix shown in Eq. \ref{attention_map2} to be the attention map $Map$ used to select the high-resolution feature that will be used in the next layer. A pseudo algorithm for how the vision-language model with our \textsc{{\methodname}} works is described in Alg. \ref{alg:flexattn}.

\begin{algorithm}[t]
    \begin{algorithmic}[1]
    \STATE \small{\textbf{Input:} High-resolution Image $I_{HR}$, Text $T$, }
    \STATE \small{\textbf{Sub-modules:} Image Encoder $E_i(\cdot)  $, Text Tokenizer $E_t(\cdot)$,  Self-Attention $A(\cdot)$, Hierarchical Self-Attention $HA(\cdot)$, Feed-Forward Network $FFN(\cdot)$, Prediction Head $Head(\cdot)$}
    \STATE \small{\textbf{Parameters:} \#Self-Attention layer $N_{SA}$, \#FlexAttention layer $N_{FA}$} \\
    \STATE Downsample $I_{HR}$ to low-resolution image $I_{LR}$.
    \STATE Generate image and text tokens $f_{HR}=E_{i}(I_{HR})$, $f_{LR}^0=E_{i}(I_{LR})$, $f_{T}^0=E_{t}(T)$
    \STATE $H^0=Concat(f_{LR}^0, f_{T}^0)$
    
    \STATE \small{\color{gray}{\# Decoder layers with self-attention}}
    \FOR{i = 1 \ldots $N_{SA}$}
        \STATE  $Map^i, H^i = A(H^{i-1})$
        \hspace{0.5cm}\small{\color{gray}{\# Self-attention}}
        \STATE $H^{i} = H^{i} + H^{i-1}$
        \hspace{1.1cm}\small{\color{gray}{\# Skip connection}}
        \STATE $H^{i} = FFN(H^{i}) + H^{i}$
        \hspace{0.35cm}\small{\color{gray}{\# FFN + skip connection}}
    \ENDFOR
    
    \STATE \small{\color{gray}{\# Decoder layers with FlexAttention}}
    
    \FOR{i = $N_{SA}$+1 \ldots $N_{SA}$+$N_{FA}$}
        \STATE $f_{SHR}^{i-1} = R(f_{HR}, Map^{i-1})$ 
        \hspace{0.85cm} \small{\color{gray}{\# Select attended high-resolution feature}}  
        \STATE $Map^i, H^{i} = HA(H^{i-1}, f_{SHR}^{i-1})$ 
        \hspace{0.3cm}\small{\color{gray}{\# Hierarchical attention}}
        \STATE $H^{i} = H^{i} + H^{i-1}$
        \hspace{2.15cm}\small{\color{gray}{\# Skip connection}}
        \STATE $H^{i} = FFN(H^{i}) + H^{i}$
        \hspace{1.45cm}\small{\color{gray}{\# FFN + skip connection}}
    \ENDFOR

    \STATE Generate output tokens from $Head(H^{N_{SA}+N_{FA}})$

    \end{algorithmic}
    \caption{\small Inference Algorithm of VLM with FlexAttention.}
    \label{alg:flexattn}
\end{algorithm}

\subsection{Complexity Analysis} \textsc{{\methodname}} offers the advantage of executing computations akin to traditional self-attention, thereby minimizing alterations to the model's architecture while facilitating an efficient fusion of multi-grained features. Let the length of the selected high-resolution feature be $M$, the length of the original hidden state be $N$, and the hidden state size be $D$. The computational complexity of our hierarchical self-attention is

\begin{align}
    \textbf{T} = \mathbf{O}((M+N)ND).
\end{align}

\noindent If not using our hierarchical self-attention and directly adding the high-resolution image along with the low-resolution one, the computational complexity will be

\begin{align}
    \textbf{T}_{original} = \mathbf{O}((M+N)^2D).
\end{align}

\noindent For vanilla self-attention, the addition of an extra high-resolution feature will lead to a quadratic increase in computation time due to the need to process a significantly larger matrix, as every additional element in the sequence adds to the computational load on a per-element basis. However, the hierarchical self-attention mechanism employed by \textsc{{\methodname}} cleverly mitigates this issue by maintaining a linear relationship in terms of the addition of high-resolution features, thereby considerably reducing the computational burden.

\section{Experiments}
\label{sec:exp}

\newcommand{\baselineHD}{{LLaVA-HD}}
\newcommand{\baselineXattn}{{LLaVA-XAttn}}
\newcommand{\ourmodel}{{LLaVA-FlexAttn}}

We evaluate \textsc{{\methodname}} on both high-resolution multimodal benchmarks \cite{wu2023vstarbench, li2023otterhd, singh2019textvqa, lobry2020rsvqa} and general multimodal benchmarks \cite{yu2016modeling, li2023pope, hudson2019gqa, antol2015vqa, liu2023mmbench, fu2023mme, yu2023mmvet}, comparing our method with the low-resolution large vision-language models \cite{dai2023instructblip, li2023otter, chen2023minigpt} as well as other high-resolution methods \cite{liu2023improved, hong2023cogagent}.

\subsection{Implementation}

To assess the performance and efficiency of our proposed \textsc{{\methodname}}, we integrated it into LLaVA-1.5-7b \cite{liu2023improved}, resulting in a variant we call {\ourmodel}. The input resolution is set to be 1008x1008, which is three times the original input image resolution. We then compared this variant with the original LLaVA-1.5-7b model to demonstrate the advantages of utilizing high-resolution image inputs. We also compare \textsc{{\methodname}} with the methods used in LLaVA-1.5-HD \cite{liu2023improved} and CogAgent \cite{hong2023cogagent} that enables the input of high-resolution image in those models, to show the efficiency of our proposed method.

\subsubsection{LLaVA-1.5-HD \cite{liu2023improved}} In this model, the high-resolution image tokens act like normal tokens. They are concatenated with the low-resolution image tokens and are fed into the large language model together. Since this model has not been publicly released yet, we re-implement it on top of the codebase for LLaVA-1.5. We use the LLaVA-1.5-7b model as the base model. The input resolution of the high-resolution image is set to 448x448 following the setting in \cite{liu2023improved}. We refer to this baseline as {\baselineHD}.

\subsubsection{CogAgent \cite{hong2023cogagent}} In this model, the high-resolution feature is perceived using a cross-attention module. In the cross-attention module, the high-resolution features serve as the key and value, while the hidden states, comprising both low-resolution image tokens and text tokens, act as the query. Since CogAgent is trained on document and GUI style data, and the data processing and training code has not been released, for fair comparison on the effectiveness of the high-resolution operator used in CogAgent, we transfer the cross-attention module in CogAgent's inference codebase to LLaVA-1.5 and re-implement the training code. We use the LLaVA-1.5-7b model as the base model. The input resolution of the high-resolution image is set to 1008x1008 to keep it the same as ours. We refer to this baseline as {\baselineXattn}.

\subsection{Training Settings}

For a fair comparison, both high-resolution baselines ({\baselineHD} and {\baselineXattn}) and our {\ourmodel} load the pre-trained weight for LLaVA-1.5-7b as initialization, and are then finetuned on the LLaVA-1.5-7b's finetuning dataset for one epoch. We use a batch size of 1152 and a learning rate of 2e-5, with a cosine learning rate scheduler. All evaluations are performed in a zero-shot manner.

\subsection{Evaluation on High-resolution Multimodal Benchmarks}
\label{sec:eval_highres_benchmark}

\textbf{Datasets.} We conduct experiments on four high-resolution benchmarks: V* Bench \cite{wu2023vstarbench}, MagnifierBench \cite{li2023otterhd}, TextVQA \cite{singh2019textvqa} and RSVQA-HRBEN \cite{lobry2020rsvqa}. The first two benchmarks focus on evaluating the model's capability on general high-resolution VQA, while the last two benchmarks focus on evaluating the model's performance on domain-specific high-resolution VQA such as TextVQA for text understanding and RSVQA-HRBEN for remote sensing.

\noindent\textbf{Baselines.} We conduct a comparative analysis between {\ourmodel} and two categories of Vision-Language Models (VLMs): low-resolution VLMs, specifically InstructBLIP \cite{dai2023instructblip}, Otter \cite{li2023otter}, MiniGPT-4 \cite{zhu2023minigpt4}, MiniGPTv2 \cite{chen2023minigpt} and LLaVA \cite{liu2023improved}, as well as high-resolution VLMs that were re-implemented for this research. Additionally, comparisons are made with commercial chatbots such as GPT-4V \cite{achiam2023gpt}, and specialist VLM such as GeoChat \cite{kuckreja2023geochat}, to evaluate the significance of high-resolution image input capabilities.

\begin{table}[t]
\centering
\resizebox{1\linewidth}{!} {
\begin{tabular}{l|c|ccc|c}
\toprule
\multicolumn{1}{c|}{} & \multirow{2}{*}{Resolution} & \multicolumn{3}{c|}{V* Bench} & \multirow{2}{*}{MagnifierBench} \\
\multicolumn{1}{c|}{}   &  & Attribute & Spatial & Overall &      \\ \midrule
\textit{Commercial Chatbots} &&&&& \\
Bard \cite{bard}                    &-& 31.3     & 46.1   & 37.2   & -  \\
Gemini Pro \cite{gemini}              &-& 40.9     & 59.2   & 48.2   &  -  \\
GPT-4V \cite{achiam2023gpt}                  &-& \textbf{51.3}     & 60.5 & \textbf{55.0} &  - \\ \midrule
\textit{Low-resolution VLMs} &&&&& \\
InstructBLIP \cite{dai2023instructblip}            & $224^2$ & 25.2     & 47.4   & 34.0   & 5.6  \\
Otter \cite{li2023otter}                   & $224^2$ & 27.0     & 56.6   & 38.7   & 25.7 \\
MiniGPT-4 \cite{zhu2023minigpt4}               & $224^2$ & 30.4 & 50.0 &  38.2 & 22.6 \\
LLaVA-1.5-7b \cite{liu2023improved}            & $336^2$ & 41.7     & 56.6   & 47.6   & 26.8 \\ \midrule
\textit{High-resolution VLMs} &&&&& \\
{\baselineHD} \cite{liu2023improved}        & $448^2$ & 45.2     & \underline{61.8}   & 51.8   & \textbf{35.0} \\
{\baselineXattn} \cite{hong2023cogagent}       & $1008^2$ & 42.6     & 56.6   & 48.2   & 32.2 \\
{\ourmodel} & $1008^2$ & \underline{47.8} & \textbf{64.5} & \underline{54.5} & \textbf{35.0} \\
\bottomrule
\end{tabular}
}
\caption{General high-resolution VQA benchmark results comparison.}
\label{tab:main-table}
\end{table}

\noindent\textbf{Results.} Table \ref{tab:main-table} shows the evaluation results on the two high-resolution general VQA benchmarks. In general, all three high-resolution VLMs are better than low-resolution VLMs, while our model is consistently better than other high-resolution VLMs, with an overall accuracy of 54.5\% for V* Bench and an accuracy of 35.0\% for MagnifierBench. Compared to the base model LLaVA-1.5-7b, the overall accuracy gain for V* Bench is 6.9\% and the accuracy gain for MagnifierBench is 8.2\%. Compared to other high-resolution methods, our method achieves comparable and even higher accuracy at the cost of much lower TFLOPs than other high-resolution methods, nearly 30\% lower TFLOPs than {\baselineHD} (from 24.9 to 17.1) and over 37\% lower TFLOPs than {\baselineXattn} (from 27.1 to 17.1). Detailed discussion on the TFLOPs and inference time can be found in Sec. \ref{infer_time_sec}. Thanks to the high-resolution feature selection and hierarchical self-attention, our method can enable the input image resolution to increase three times compared to the original resolution, with the cost of a sub-linear computational cost increasing, achieving a better trade-off between computational cost and accuracy. Compared with GPT-4V on V* Bench, our method shows competitive performance, achieving even higher accuracy on spatial category than GPT-4V, and a comparable overall performance with GPT-4V.

\begin{table}[t]
\centering
\begin{tabular}{l|ccc|c}
\toprule
 & \multicolumn{3}{c|}{RSVQA-HRBEN} & \multirow{2}{*}{TextVQA} \\
 & Presence & Comparison & Overall &  \\ \midrule
\textit{Low-resolution VLMs} &&&& \\
GeoChat \cite{kuckreja2023geochat} & 58.5 & 83.2 & 72.3 & -\\
MiniGPTv2 \cite{chen2023minigpt} & 40.8 & 50.9 & 46.5 & 27.5 \\
LLaVA-1.5-7b \cite{liu2023improved} & 69.8 & 67.3 & 68.4 & 46.0 \\ \midrule
\textit{High-resolution VLMs} &&&& \\
{\baselineHD} \cite{liu2023improved} & 69.0 & 67.6 & 68.4 & 45.6 \\
{\baselineXattn} \cite{hong2023cogagent} & 71.4 & 70.9 & 71.1 & 45.5 \\
{\ourmodel} & \textbf{72.2} & \textbf{73.1} & \textbf{72.7} & \textbf{48.9}\\
\bottomrule
\end{tabular}
\caption{Domain-specific high-resolution VQA benchmark results comparison.}
\label{tab:domain-specific-table}
\end{table}

Table \ref{tab:domain-specific-table} presents the results on two high-resolution domain-specific VQA benchmarks. Our {\ourmodel} is consistently superior to the base model and other high-resolution methods on both RSVQA-HRBEN and TextVQA. Furthermore, our approach surpasses GeoChat \cite{kuckreja2023geochat} in terms of overall accuracy on the RSVQA-HRBEN benchmark, a model explicitly crafted and fine-tuned for remote sensing Visual Question Answering benchmarks. This outcome underscores the efficacy of incorporating high-resolution image inputs, suggesting that the increased detail and clarity provided by high-resolution inputs can significantly improve the model's understanding and processing of intricate visual patterns in specialized VQA tasks.

\subsection{Evaluation on General Multimodal Benchmarks}
\label{sec:gen_ben}
\subsubsection{Datasets and Baseline.} We evaluate the general vision-language model performance on several multimodal tasks including GQA \cite{hudson2019gqa}, VQAv2 \cite{antol2015vqa}, POPE \cite{li2023pope}, RefCOCO \cite{yu2016modeling}, MM-Bench \cite{liu2023mmbench}, MME \cite{fu2023mme}, and MM-Vet \cite{yu2023mmvet}. This collection of benchmarks assesses the model's overall capabilities, including spatial understanding, localization, ability to avoid hallucinations, and performance in academic-oriented tasks. We compare our method to the base model LLaVA-1.5-7b to analyze the change in the model's general ability.

\begin{table}[t]
\centering
\begin{tabular}{l|cccccccc}
\toprule
              & RefCOCO &  POPE & GQA  & VQAv2 & MM-Bench & MME & MM-Vet \\ \midrule
LLaVA-1.5-7b \cite{liu2023improved}  &  75.8 &  85.9 & 62.0 & 78.5  & 64.3  & \textbf{1511} & \textbf{31.1}   \\
{\ourmodel} & \textbf{79.3} &  \textbf{85.9} & \textbf{62.2} & \textbf{78.7}  & \textbf{65.7} & 1479 & 29.4 \\
\bottomrule
\end{tabular}
\caption{Comparison of the multimodal capability between the base model and our model on a broad range of multimodal benchmarks.}
\label{tab:general_vqa}
\end{table}

\noindent\textbf{Results.} In Table \ref{tab:general_vqa} we show that with our \textsc{{\methodname}}, the performance on RefCOCO is improved. RefCOCO requires the localization of an object based on a referring expression. Thus, incorporating a high-resolution feature could reduce the challenge of identifying a small object and enhance the precision of its location prediction. We achieve a similar rate of hallucination on POPE and maintain similar performance on large-scale VQA benchmarks. This indicates that incorporating \textsc{{\methodname}} does not impact the model's overall capability.

\subsection{Ablation Study}

\textbf{H.R. Feature Selection Strategy.} We first conduct an ablation study to verify the effectiveness of the key design of our method, which is the strategy to select high-resolution features using the attention map. We compare our attention map selection strategy with two naive baseline strategies: 1) random selection, which means randomly selecting a few patches of the high-resolution features, and 2) center selection, which means selecting the center region of the high-resolution features. The selection ratio is kept to approximately the same as our attention map selection strategy which is about 10\%. We finetune them respectively using the same finetuning dataset and following the same training setting and evaluate their performance on Magnifierbench and TextVQA.

\begin{figure}[t]
\centering
\begin{minipage}[c]{0.47\textwidth}

\begin{table}[H]
\centering
\resizebox{1\linewidth}{!}{
\begin{tabular}{l|cccc}
\toprule
 & Magnifierbench & TextVQA \\ \midrule
Random & 31.4 & 44.5  \\
Center & 30.7 & 45.9 \\
Attn. Map & \textbf{35.0} & \textbf{48.9} \\
\bottomrule
\end{tabular}
}
\hfill
\centering
\end{table}

\end{minipage}
\begin{minipage}[c]{0.47\textwidth}
\includegraphics[width=0.95\linewidth]{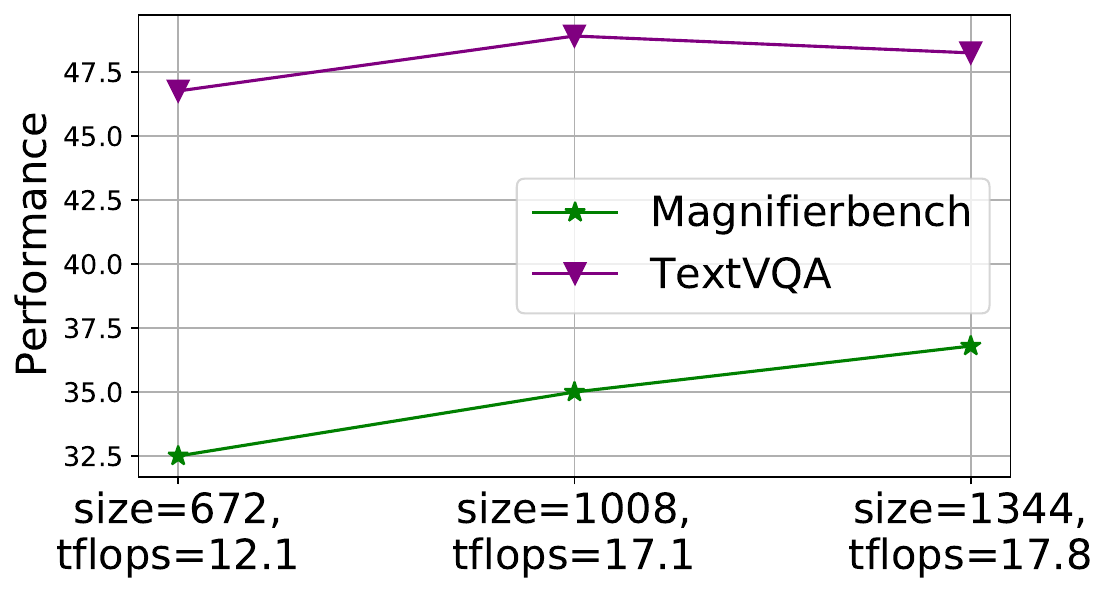}
\centering
\end{minipage}
\caption{Ablation studies of selection strategies (left) and image sizes (right).}
\label{fig:abs}
\end{figure}

\begin{table}[t]
\centering
\footnotesize
\begin{tabular}{l|ccc}
\toprule
RefCOCO Val Acc. & Large       & Small        & Overall     \\ \midrule
LLaVA            & 75.9        & 41.3         & 75.4        \\
LLaVA-FlexAttn   & 78.8 (+2.9) & 51.3 \textbf{(+10.0)} & 78.4 (+3.0) \\
\bottomrule
\end{tabular}
\caption{Analysis on accuracy across different object sizes.}
\label{tab:refcoco_acc}
\end{table}

The experiment results in Fig. \ref{fig:abs} (left) shows that our attention map selection strategy is better than the other two baseline strategies. For the baseline strategies, since the model cannot dynamically pay more attention to the region that needs to be focused, the benefit of the high-resolution image is limited, and no consistent improvement is observed especially for TextVQA which requires the model to focus on a specific region to give the correct answer.

\noindent\textbf{Impact of Resolution.} We also explore the effect of the high-resolution image size. The default setting is 1008x1008, tripling the resolution of the original low-resolution image. Additionally, we introduce two other settings: 672x672 and 1344x1344, doubling and quadrupling the original resolution, respectively. We finetune them respectively using the same finetuning dataset and following the same training setting. We measure their average TFLOPs on Magnifierbench benchmarks and evaluate their performance on Magnifierbench and TextVQA.

Fig. \ref{fig:abs} (right) shows the experiment results. We can see that as the resolution increases, the performance on Magnifierbench also increases. Performance on TextVQA significantly enhances when the resolution is increased from 672 to 1008 but sees no further improvement from 1008 to 1344. Since the average resolution of images in TextVQA is $950\times811$, further increasing the resolution beyond its original resolution is unbeneficial. This pattern is aligned with what we observe for general VQA benchmarks.

\noindent\textbf{Impact of Object Size.} For general benchmarks evaluated in Section~\ref{sec:gen_ben}, most questions do not focus on small details, and thus cannot reveal the capability of our model for handling high-resolution image. To better evaluate our model on general benchmarks, we divide the benchmark into two subsets according to the size of question-relevant objects, categorizing those larger than 5\% of the image as large objects and the rest as small. We conduct experiments on RefCOCO val set as it provides object sizes.

Table~\ref{tab:refcoco_acc} shows that the accuracy improvement on small objects is much higher than large objects. It indicates that even for non high-resolution benchmarks such as RefCOCO, our method can still improve the accuracy when the questions involve small objects or detailed information. 

\subsection{Inference Time on Hardware}
\label{infer_time_sec}

\noindent We measure the inference time on hardware to assess the efficiency of our \textsc{{\methodname}}. Models are implemented in PyTorch and the inference time is measured on a single NVIDIA V100 32G GPU. We measure the average TFLOPs and total inference time on two benchmarks: Magnifierbench, in which the model answer is a single letter, and TextVQA, in which the model answer is a short phrase. Warm-up before inference and CUDA synchronization are employed to ensure the accuracy of the measurement results.

The measurement results are presented in Table \ref{tab:infer_time}. In Magnifierbench, the inference time reduction is linearly proportional to the theoretical computational cost reduction measure in TFLOPs, and our method is nearly 30\% and 40\% faster than the two baselines respectively. In TextVQA, the speed superior slightly declined, but still about 15\% and 25\% faster than baselines. Note that the average output length for TextVQA is longer than Magnifierbench, so the inference time will be affected more by the generation phase, which is memory-bound instead of computation-bound. A discussion is provided in the Supplementary.

\begin{table}[th]
\centering
\begin{tabular}{l|cc|cc}
\toprule
 & \multicolumn{2}{c|}{Magnifierbench} & \multicolumn{2}{c}{TextVQA} \\
 & TFLOPs & Time(s) & TFLOPs & Time(s) \\ \midrule
{\baselineHD} \cite{liu2023improved} & 24.9 & 154 & 24.5 & 3273  \\
{\baselineXattn} \cite{hong2023cogagent} & 27.1 & 178 & 26.7 & 3741 \\
{\ourmodel} & \textbf{17.1} & \textbf{112} & \textbf{17.1} & \textbf{2839} \\
\bottomrule
\end{tabular}
\caption{Average TFLOPs and total inference time measured on NVIDIA V100 GPU.}
\label{tab:infer_time}
\end{table}
\section{Conclusion}

In this paper, we propose \textsc{{\methodname}}, a method designed to enhance large vision-language models by allowing them to efficiently process and derive advantages from high-resolution image inputs. By leveraging dynamic high-resolution feature selection and hierarchical self-attention mechanism, \textsc{{\methodname}} surpasses existing high-resolution methods in terms of performance as well as efficiency. The idea behind \textsc{{\methodname}} can be extended to other long sequence modalities such as video or audio, which can be a crucial future direction.

\section*{Acknowledgments}

We are grateful to the anonymous reviewers for their valuable feedback. We also extend our thanks to AiMOS for supplying the computational resources necessary for this project.


%
%
\bibliographystyle{splncs04}
\bibliography{main}
\end{document}